\documentclass{llncs2e/llncs}
\usepackage{epsfig}

\title{The Impact of Social Networks on Multi-Agent Recommender Systems}
\author{Hamilton Link$^{1,2}$ \and Jared Saia$^1$ \and Terran Lane$^1$ \and Randall A. La{V}iolette$^2$}
\institute{Department of Computer Science \\ University of New Mexico \\ Albuquerque, NM
  \and Sandia National Laboratories \\ Albuquerque, NM }
\date{\today}

\begin{document}

\maketitle

\begin{abstract}
Awerbuch et al.'s approach to distributed recommender systems (DRSs)
is to have agents sample products at random while randomly querying
one another for the best item they have found; we improve upon this by
adding a communication network.  Agents can only communicate with
their immediate neighbors in the network, but neighboring agents may
or may not represent users with common interests.  We define two
network structures: in the ``mailing-list model,'' agents representing
similar users form cliques, while in the ``word-of-mouth model'' the
agents are distributed randomly in a scale-free network (SFN).  In
both models, agents tell their neighbors about satisfactory products
as they are found.  In the word-of-mouth model, knowledge of items
propagates only through interested agents, and the SFN parameters
affect the system's performance.  We include a summary of our new
results on the character and parameters of random subgraphs of SFNs,
in particular SFNs with power-law degree distributions down to minimum
degree 1.  These networks are not as resilient as Cohen et
al.\ originally suggested.  In the case of the widely-cited ``Internet
resilience'' result, high failure rates actually lead to the orphaning
of half of the surviving nodes after 60\% of the network has failed
and the complete disintegration of the network at 90\%.  We show that
given an appropriate network, the communication network reduces the
number of sampled items, the number of messages sent, and the amount
of ``spam.''  We conclude that in many cases DRSs will be useful for
sharing information in a multi-agent learning system.
\end{abstract}

\section{Introduction}

One of the canonical problems of machine learning is recommending
products to potential users, i.e., presenting each user with one or
more items they are likely to be satisfied with.  Recommendations can
be given based on features of products, user similarity, or both, but
the algorithms that learn from this data are typically centralized
\cite{basu98recommendation,melville02contentboosted}.  Awerbuch et
al.\ \cite{awerbuch05improved} presents what appears to be the first
distributed recommendation system (DRS) algorithm, albeit with a
relaxed definition; the goal is to ensure that most users are
\begin{em}eventually\end{em} presented with a satisfactory item.
Even so, distributed recommender systems potentially have many
attractive qualities, such as privacy and fault-tolerance.  They avoid
gathering personal information from a large number of users back to a
central repository.  A large system could learn how best to distribute
announcements of new technical papers, for example, by locally
capturing information on research interests and sharing key pieces of
information only between immediate peers. Such a system is also more
robust to failures than a centralized one, and it does not require
corporate interest to ensure servers were maintained (there being only
a peer-to-peer network or similar infrastructure to begin with).  If
the distributed algorithm could be designed with a suitable
abstraction barrier between nodes, personalization may also be
enabled, with each user's system independently learning the structure
and parameters of that user's interests and how each neighbor's
preferences are related.

We consider the impact of an explicit communication network in a DRS
with requirements similar to \cite{awerbuch05improved}.  For this
work, agents (the DRS nodes that act on each user's behalf) were
placed into a graph, and were only able to communicate with their
immediate neighbors.  Random polling was dropped in favor of local
broadcast.\footnote{This would have been untenable in the model of
  \cite{awerbuch05improved}, which uses global interactions.}  Given
this new model, we considered the algorithm's performance under two
graph structures: one in which users with substantial common interests
have agents organized into cliques, and one in which the agents are
randomly connected in a scale-free network \cite{aiello00random}.
With a network, the amount of work required to sample the space of
products is improved, in part because broadcasts are used to spread
knowledge of liked items more rapidly.  The communication complexity
is also greatly improved, and this includes a reduction in the amount
of uninformative traffic, or ``spam.''

In the mailing-list model, agents with significant common interests
are connected to one another by design. Such a system must be
engineered to ensure that new members are properly connected.  It is
conceivable, however, that other network structures would naturally
ensure that new members would serendipitously connect to at least one
existing member with common interests, effectively causing groups of
users with overlapping interests to form connected subgraphs in the
population.  Previously published results by Cohen et al.\ on the
resilience of scale-free networks to random node failures
\cite{cohen00resilience} suggested SFNs might have this property, and
this led us to use SFNs for the word-of-mouth model.  Although the
original resilience results do not seem to apply to SFNs of minimum
degree 1, SFN structure varies with the power-law exponent used, and
we have identified parameters for which a DRS would perform well.

We have a more accurate estimation of the properties of scale-free
networks when a potentially large fraction of the nodes fail uniformly
at random.  The approximations used in the original work on this
subject \cite{cohen00resilience} led to the claim that the
Internet\footnote{The Internet is cited as a scale-free network with
  an exponent of 2.5 and minimum degree 1 in \cite{cohen00resilience},
  but see also \cite{achlioptas05bias,chang04towards,newman03mixing}.}
would retain a spanning cluster across surviving nodes even if 99\% of
its nodes failed.  We conclude, based on more exact formulation, that
the original approximations are highly optimistic when the degree
distribution obeys a power law down to a minimum degree of 1, and that
the critical fraction is at best 89.8\%.  Furthermore, under such
conditions any ``spanning'' component would only capture a small
constant fraction of the surviving nodes -- if 60\% or more of the
network was to fail, at least half of the remaining nodes would be
neighborless orphans.  We achieved this result by specifically
considering the degree-1 case, explicitly considering the portion of
the network that is orphaned, and approximating the resulting network
with explicit size and power-law parameters as in
\cite{aiello00random}.

Section~\ref{section-relatedWork} reviews the most relevant literature
in recommender systems and cites the most relevant literature on
scale-free networks; section~\ref{section-DRSs} covers the distributed
recommender system model from \cite{awerbuch05improved} in more detail
and introduces notation that will be used
later. Section~\ref{section-SFNs} outlines our result in the
conditions for resilience of scale-free networks and explains its
relevance to DRSs.  Sections \ref{section-MLM} and \ref{section-WMM}
use the established notation and graph properties to derive our
algorithm's performance in the number of sampled items, the number of
messages sent, and the amount of ``spam'' sent. The final section
summarizes our results and the work that remains to be done.

\section{Related Work}\label{section-relatedWork}

The Internet is full of examples of centralized recommender systems
(Google, Amazon, etc.). These systems are a mix of collaborative
filters and content-based recommendation systems, but these systems
all collect, analyze, and use the information centrally. This is
perfectly legitimate, but it is not the only available strategy; in
many cases the recommendations given are either the service being
provided or are a customer service related to the hosting business.
Centralized algorithms for recommender systems have been covered
extensively in the literature
\cite{adomavicius05toward,mladenic99textlearning}.

Distributed recommendation systems are not as highly represented.  In
\cite{awerbuch05improved} the definition of ``recommendation system''
is altered slightly to require that users eventually find a
satisfactory item in a stream of recommendations, instead of being
given a short list of the most promising items.  Conceptually, we
would like to design systems like \cite{awerbuch05improved} that use a
DRS to efficiently pass information about potentially desirable items
through to potential users and then locally apply a user-specific
centralized content-based algorithm to filter and rank items as they
arrive.  This is like an idealized email system in which people only
send messages within their social circle, and in which all users have
a learning mail filter such as PopFile \cite{graham-cumming02popfile}.
From the centralized point of view the most analogous work we have
found is content-boosted collaborative filtering (CBCF)
\cite{melville02contentboosted}.

Scale-free networks were originally of interest to us because of their
published resilience to random failures
\cite{cohen00resilience,saroiu02measurement}, which implied random
subgraphs of an SFN had a good chance of being highly connected
(subject to the power-law exponent of the original graph).  This
suggests that simply propagating items through the network from their
point of discovery, in a manner similar to \cite{awerbuch05improved},
would with high probability reach most or all of the agents that would
be interested in the item.  Recent SFN literature can be roughly
divided into theoretical and empirical camps. The formal treatments
are based in physics, statistical mechanics, and mathematics
\cite{aiello00random,albert02statistical,cohen00resilience,molloy98size}
and describe the mathematical properties that can be derived from the
assumption that SFN node degrees follow a power-law distribution.  The
study-driven work
\cite{csanyi04structure,reed04brief,saroiu02measurement} is aimed at
capturing or sampling the structure and degree distribution of
real-world networks such as electric power transmission, Internet
routing, web pages, social networks, etc.\ in order to see if the
observed systems are scale free, and to verify the theoretical
properties.  Many authors refer to the tendency of non-engineered
Internet communities to form scale-free networks, although for the
Internet some of this work has used potentially biased forms of
sampling \cite{achlioptas05bias}.

The viral spread of information (or pathogens) in social and
information networks has also been studied extensively
\cite{kempe03maximizing}, and the relationship between recommender
systems and epidemic spread in subpopulations of an SFN is of interest
for further study.  Another closely related subdomain of graph theory
focuses on ``small-world'' networks \cite{watts98collective}.  We have
not analyzed a DRS overlaid on these graphs but this would be an
obvious possible extension to our work.

\section{Distributed Recommender Systems}\label{section-DRSs}

In \cite{awerbuch05improved} users search for items they like in some
set of products. They alternate between sampling the set of all
products and asking other users for recommendations.  The
\begin{em}raison d'\^{e}tre\end{em} of these users is to try products
in $P$, find things they like, and send messages about good items to
the other members of their ``special interest group'' (SIG).  In a
slightly less abstract world, our users have software agents acting on
their behalf in a network.  The agents must learn their user's
preferences, search $P$ and solicit their user's opinion on items, and
forward messages to one another.

Assume we have a set of agents $U$ of size $\mu$, and a set of
products $P$ of size $\eta$, with each user $u \in U$ having a
predetermined but unknown set of products they will like, $P(u)$.  It
is reasonable to assume that SIG members have more in common with one
another than a member's fraction of desirable products in $P$,
i.e. $\frac{P(u)}{\eta} < \frac{P(S)}{P(v)}$. If this were not the
case, items recommended by other users would have a lower expectation
of satisfying the recipient than something chosen at random.  Let
there be a set of special interest groups ${\cal{S}} = \{S : S \subset
U,\ \bigcap_{u \in S} P(u) \not= \phi\}$ , where each SIG $S$ is a set
of users with common interests.  As in \cite{awerbuch05improved}, the
stated goal is that ``a large fraction $\lambda$ of the users will
find a good product in the set recommended to them'' given that
``there exists a small collection of SIGs that cover most users,''
i.e. the fraction of $U$ represented by all of ${\cal{S}}$ is at least
$\lambda$ and the number of SIGs $\ell = |{\cal{S}}|$ is of order
$\Theta(1)$.  In \cite{awerbuch05improved} there is no requirement
that the set of products recommended to an individual be small, but it
should be clear that this would be desirable.

\begin{figure}
\label{mlm-wmm-graph}
\centerline{
  \begin{tabular}{cc}
  \epsfxsize=2in \epsfbox{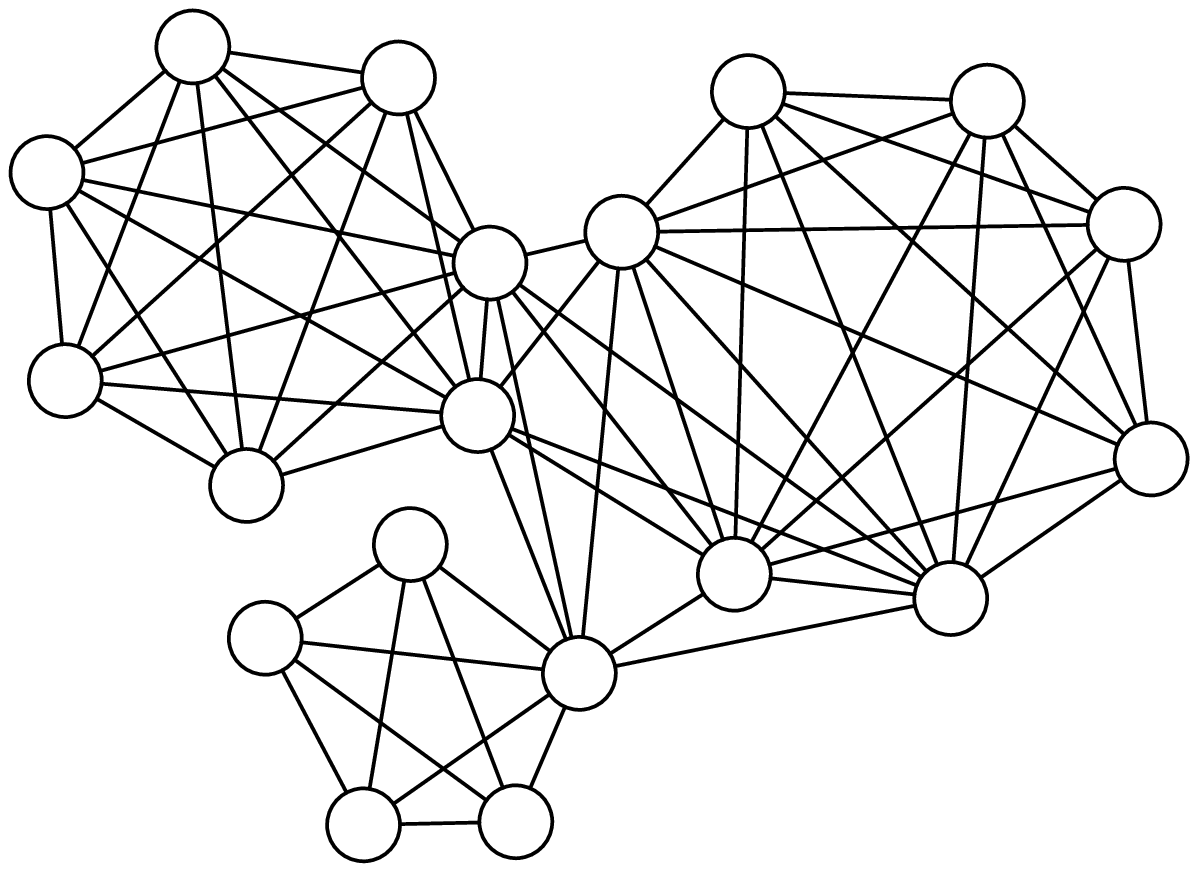} &
  \raisebox{0.15in}{\epsfxsize=2in \epsfbox{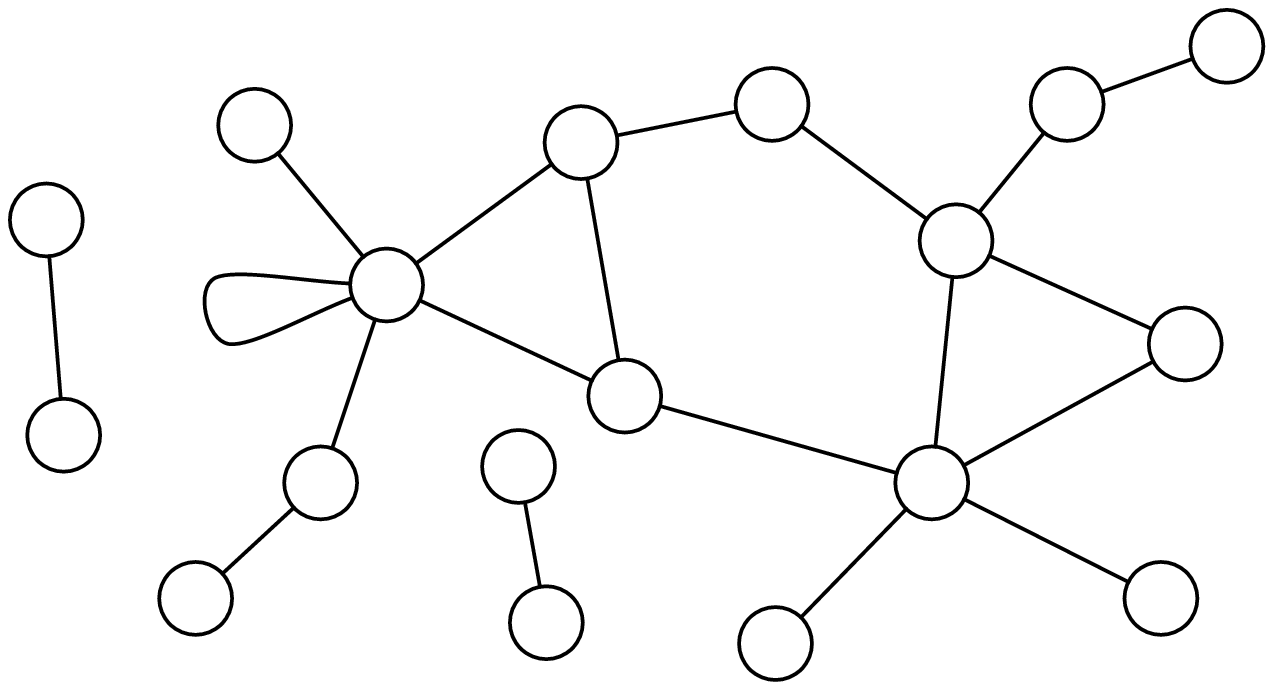}} \\
  (\textbf{a}) & (\textbf{b})
  \end{tabular}}
\caption{The mailing-list model (a) with four SIGs sizes $5$, $7$,
  $7$, and $6$, and the word of mouth model (b) generated with a
  random configuration given a power-law degree distribution with
  $\alpha = 2.3$ and $\beta = 1.2$.  Notice the presence of a self-arc
  in the latter.  Self-arcs are an inevitable artifact of generating
  simulated networks using the random configuration model but occur in
  no significant fraction in large graphs.}
\end{figure}

For this work, we add a graph $G$ representing the communication
network available to the users.  In the ``mailing-list model,'' users
are organized into (potentially overlapping) cliques based on their
SIGs (Figure~\ref{mlm-wmm-graph}a).  In the ``word-of-mouth model,''
the users are randomly distributed in a scale-free network
(Figure~\ref{mlm-wmm-graph}b). In either model, we can consider the
subgraph $G'$ comprising some set of vertices in $G$ (corresponding to
the members of a SIG).  Users communicate by broadcasting positive
findings to their immediate neighbors in the graph.  When a user $u$
receives a product name from $v$, they sample the item, and may or may
not generate further messages (depending on the structure of the
network) if it is satisfactory. For our analysis, users sample shared
items independently of their continuous random sampling.

A related question to the amount of work required to disseminate
popular items (and the portion of this that leads to unfruitful probes
of products, if probes have an associated cost) is the amount of
unproductive traffic, or ``spam'', generated during the cooperative
process. In the mailing-list model and word-of-mouth model, spam
consists of an item being reported to any individual for whom that
item is not in their set of interests.  For the algorithm in
\cite{awerbuch05improved} we consider spam to be any request for a
recommendation sent to an unsatisfied individual, although this
definition could be extended to include other unproductive traffic.

It is important to stress that this decentralized system is not
solving the same problem as traditional recommendation systems; this
system is not identifying or exploiting similarities between users or
trying to estimate the likelihood that some user will like some item.
Instead, the goal is to ensure that information about some item will
\begin{em}reach\end{em} most of the users that are likely to be
satisfied by the item.  Then, a content-based algorithm would be used
to estimate how desirable the item is for each the users it reaches.
CBCF algorithms are a solution of interest, if we are able to
decentralize the learning problem and exploit existing social
structures along with domain information.

\section{Scale-Free Networks}\label{section-SFNs}

A frequently cited result in scale-free networks is the incredible
resilience to random failures of ``the Internet,'' an example of an
SFN with fairly low minimum degrees \cite{cohen00resilience}.  A
corollary of that result is that small random subgraphs of such an SFN
would exhibit a giant component.  This would be useful in a DRS
setting such as the word-of-mouth model: disinterest in an item may be
seen as node failure, and the persistence of a giant component would
imply the remaining SIG is able to pass information.  Unfortunately,
the resilience result is based on approximations that do not appear to
hold when the minimum degree of the SFN is 1 (as in the Internet).  To
answer performance questions related to DRSs in SFNs of minimum degree
1, we determined the conditions under which random SFN subgraphs remain
connected.  The full derivation can be found in draft form as
\cite{link05parameters}.

\begin{figure}
\label{sigSizeVsExp}
\centerline{
  \makebox[0pt]{\epsfxsize=3.25in \epsfbox{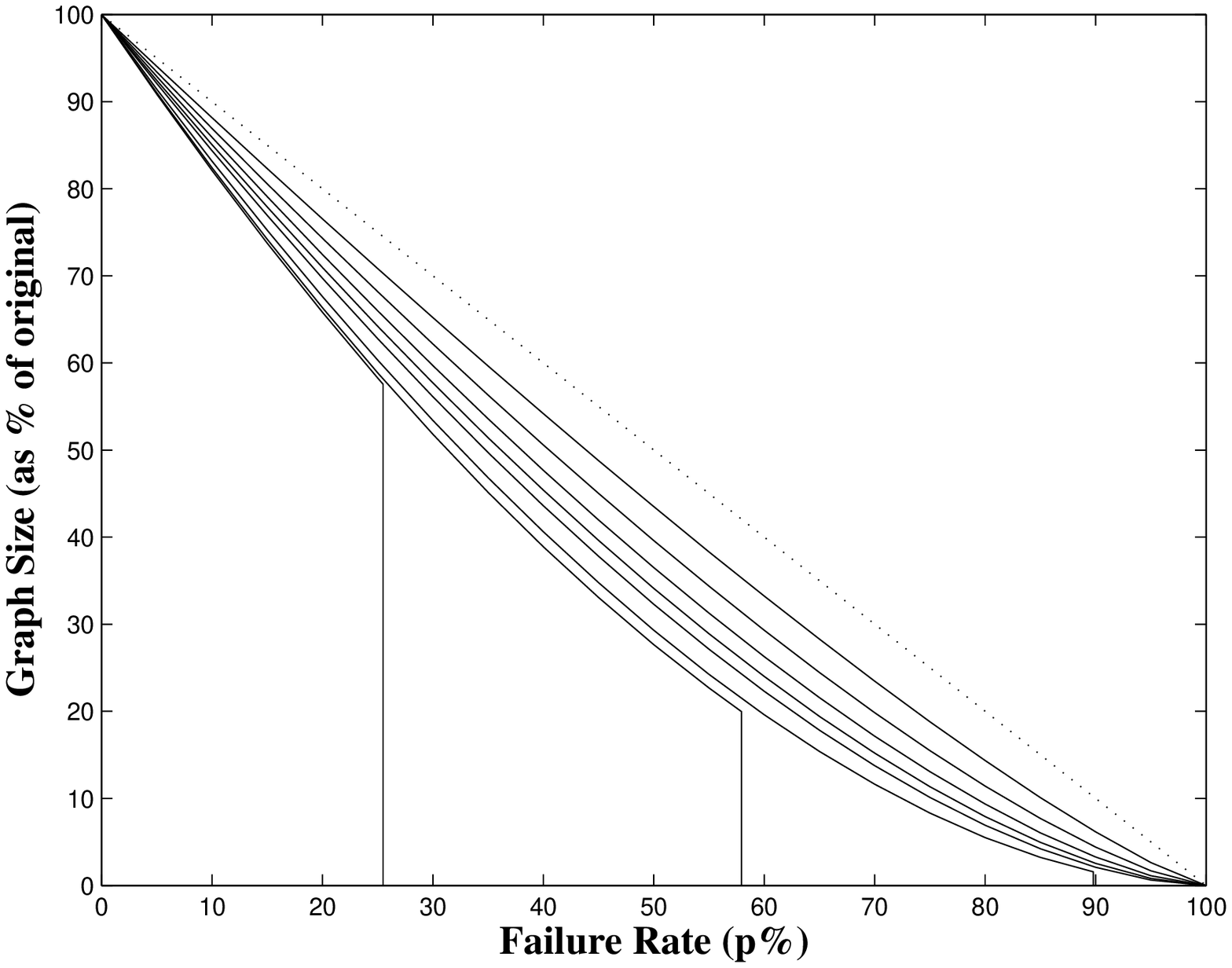}}
  \makebox[0pt]{\makebox[2.95in][r]{\raisebox{1.35in}{\epsfxsize=1.4in \epsfbox{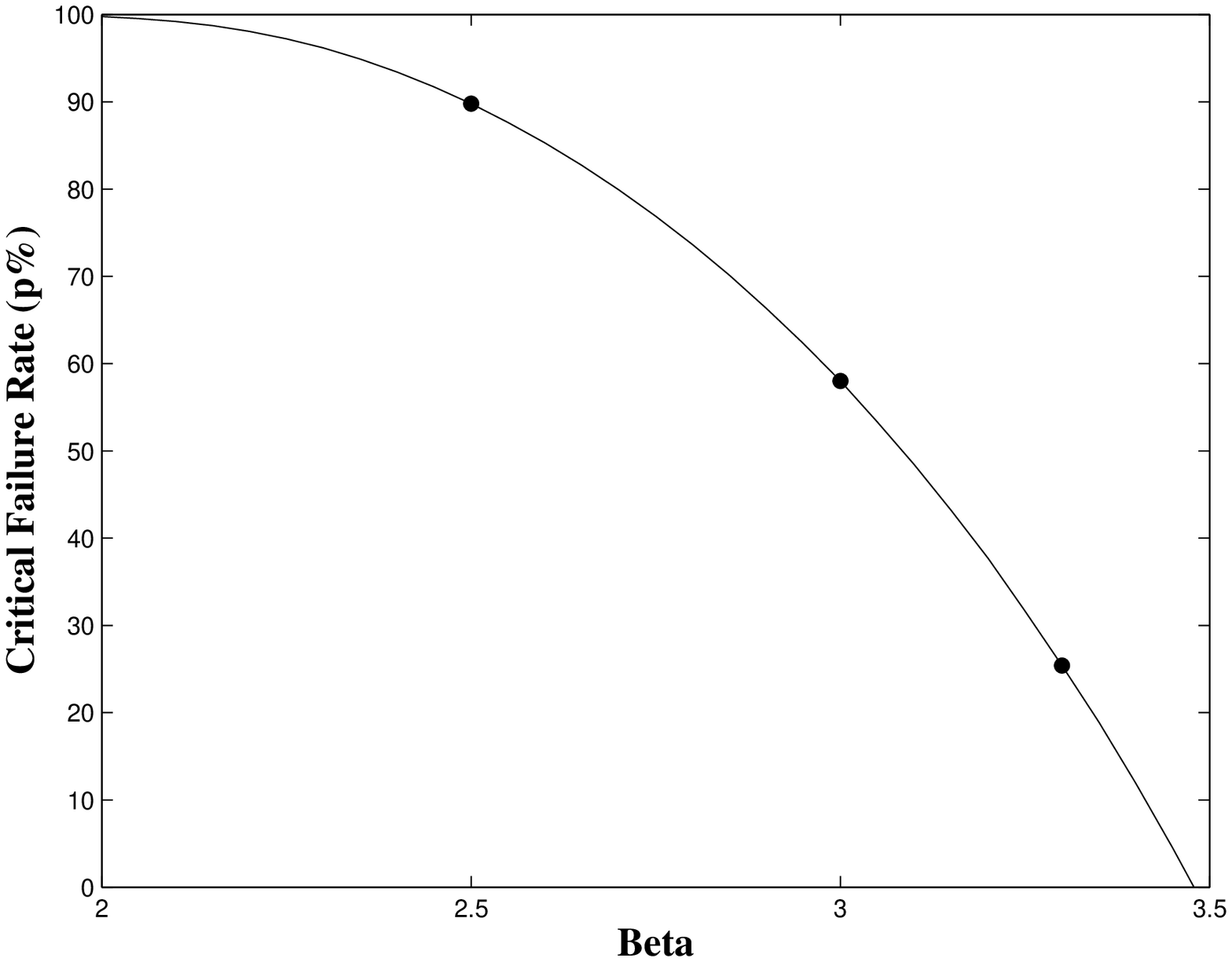}}}}
}
\caption{The size of a SIG's subgraph (minus orphans), relative to its
  host SFN, as a function of the failure rate $p$.  The distance
  between each curve and the diagonal $(1-p)$ equals the fraction of
  orphans $\#'(0)$.  Curves are for $\beta = 1.2$, $1.4$, $1.6$,
  $1.8$, $2.0$, $2.5$, $3.0$, and $3.3$.  When $2 < \beta < \beta_0$
  there is a significant critical failure rate (inset) due to the
  increase in $\beta'$; the critical points for the curves of $\beta =
  2.5$, $3.0$, and $3.3$ are highlighted.  For $\beta' < 2$ virtually
  all of the remaining graph is in the giant cluster, and this
  fraction declines from 90\% to virtually nothing as $\beta'$ goes
  from $2$ to $\beta_0$}
\end{figure}

Random subpopulations of a scale-free network have a degree
distribution that can be roughly estimated with a power law. In fact,
the distribution starts with a minimum degree of 1 and is very nearly
log/log linear, but with a rolloff that underrepresents high-degree
nodes. The rolloff causes potential giant components in an SFN to
disintegrate more readily, as shown in
\cite{cohen01breakdown,dorogovtsev01comment}, so using a pure power
law in our derivations leads to an upper bound result on the critical
failure rate $p$ -- where $(1-p)$ is the percentage size of the
subpopulation of interest -- at which point the giant component ceases
to exist, given the initial graph's power law parameter $\beta$
(Figure~\ref{sigSizeVsExp}).  This gives a lower bound on the size of
SIGs necessary in the word-of-mouth model given $\beta$; our analysis
also shows the relationship of $\beta$ and $p$ to the portion
$\lambda$ of the population that must eventually be satisfied in the
DRS problem.

The number of nodes that retain no neighbors and the the number of
nodes retaining only one neighbor are, respectively,
\begin{eqnarray}
\#'(0) = (1-p) e^\alpha \chi\mbox{,} \quad & \chi \stackrel{\mbox{\tiny def}}{=} \sum_{k_0 = 1}^{e^{\frac{\alpha}{\beta}}} \frac{1}{k_0^\beta} p^{k_0}\mbox{, and} \nonumber \\
\#'(1) = (1-p) e^\alpha \xi\mbox{,} \quad & \xi \stackrel{\mbox{\tiny def}}{=} \sum_{k_0 = 1}^{e^{\frac{\alpha}{\beta}}} \frac{1}{k_0^\beta} k_0 (1-p) p^{k_0 - 1}\mbox{.} \nonumber
\end{eqnarray}

For networks with strict power-law degree distributions and minimum
degree 1, there is no spanning component in the remaining graph when
the slope increases beyond $\beta_0 = 3.47875$ \cite{aiello00random}.
The key result is that the slope of the degree distribution of the
subgraph is greater than the slope of its parent's -- putting it
closer to or even beyond this threshold.  The new slope $\beta'$ is a
function of the original slope $\beta$ and the failure rate $p$
(captured in the sums $\xi$ and $\chi$ above):
$$\beta' = \zeta^{-1}\left( \frac{\zeta(\beta) - \chi}{\xi} \right) \mbox{.}$$

By varying $\beta$ in an engineered network, we can ensure random
subgraphs of a given fractional size will be highly connected.  In
this way, $\beta$ can be used to design a DRS network where SIGs of a
given size may propagate recommendations of particular interest.  In
contrast, for typical values of $\beta$ in empirically studied SFNs
such as the Internet, a SIG comprising a small randomly-distributed
group of DRS members has little chance of being able to share
information without the assistance of agents outside the SIG.  In
addition, when the properties of the network are known and fixed, this
result can be used to control the likelihood with which agents forward
uninteresting recommendations, artificially boosting the effective SIG
and ensuring that disconnected agents with common interests can still
reach one another.

\section{The Mailing-List Model}\label{section-MLM}

Awerbuch's model provides a fundamental theoretical result
\cite{awerbuch05improved}, but in practical terms people operate
within the auspices of social circles that can provide a more
structured (but still distributed) approach to finding items of
interest. One possible model is that users interested in certain kinds
of items will subscribe to ``mailing lists'' so that they can share
findings with one another deliberately.  While still oversimplified,
this model gives additional structure and lends itself to some initial
analysis.  Given the users and their SIGs, we can construct a graph of
user associations.  In the ``mailing-list model'' each SIG is
represented by a clique.  Let $G$ be the graph $(U, E)$ such that
$\forall\ S \in {\cal{S}}, \forall\ u,v \in S, (u, v) \in E$.  When a
user broadcasts a finding to all his neighbors, this is tantamount to
sending an email to all of the mailing lists to which they subscribe;
given the construction of $G$ this will reach the members of all of a
user's SIGs, so no retransmissions are necessary.

As in \cite{awerbuch05improved}, consider a sequence of samples,
$\sigma = \pi_1 \pi_2 ... \pi_n$, made by the members of a lone SIG
$S$. The expected number of samples $\mbox{E}[n]$ before all SIG
members are satisfied is at worst the number of messages generated
before they find an item in $P(S)$. This is equivalent to a sequence
of Bernoulli trials, giving $\mbox{E}[n] = \eta / |P(S)|$, with each
SIG member making $\frac{\eta}{|P(S)| |S|}$ samples.  For the last
element of the sequence it takes $|S|$ messages to broadcast the
finding, but some number of messages that do not satisfy the entire
SIG may also be announced in the process. Each user's samples prior to
this are either in $P(u) \!\! \setminus \!\! P(S)$ or $P \!\!
\setminus \!\! P(u)$, with only the former generating a broadcast.  In
a sequence of Poisson trials with each user $u$ having a potentially
different chance $|P(u)| / \eta$ to find an item they like in a single
trial, the expected number of liked items found will be tightly
bounded around $\frac{\mbox{\scriptsize{avg}}_{u \in S}{|P(u)|}}{\eta}
\frac{\eta}{|P(S)|} = \frac{\mbox{\scriptsize{avg}}_{u \in
    S}|P(u)|}{|P(S)|}$.  Given this, the sample
complexity\footnote{\cite{awerbuch05improved} defines recommendation
  complexity as the total number of times users test recommended
  products.  We use sample complexity for the number of time users
  test products, and reserve recommendation complexity for the number
  of times users test \begin{em}recommended\end{em} products.} of the
system is at most $$C =
\ell\left(\frac{\eta}{\mbox{\scriptsize{min}}_i\{|P(S_i)|\}} +
\mbox{avg}_i\!\left\{ |S_i|\frac{\mbox{\scriptsize{avg}}_{u \in
    S_i}|P(u)|}{|P(S_i)|}\right\} \right)\mbox{.}$$ This captures the
samples taken by each SIG to find some liked item, and the number of
extra samples users draw based on posted items.  When users belong to
more than one clique individual SIG interests $P(S)$ can be expected
to vary in size, and the sequence $\sigma$ to vary in length as a
result, but the number of samples an individual user takes is simply
the appropriate number for the worst-case SIG. This is because all
users are broadcasting into all their SIGs, with each SIG deciding
somewhat independently whether the item is of interest.  Individual
users' interests remain constant through this process and so the
number of broadcasts they are expected to make also remains the same
from the point of view of each SIG.

The mailing-list model improves upon the distributed algorithm in
\cite{awerbuch05improved}, which has a sample complexity of
$2\ell\left(\frac{\eta}{\mbox{\scriptsize{min}}_i\{|P(S_i)|\}} +
\mu\ln(\mbox{max}_i\{|U(S_i)|\})\right)$, and can be approximated as
$O(\ell(\eta + \mu \ln \mu))$.  In comparison $C = O(\ell(\eta +
\mu))$ with the additional savings of the hidden constant factor of
$2$.  The savings in samples required caused by the elimination of the
original algorithm's sample/query alternation, which causes samples to
be taken long after a suitable item for a SIG has been discovered, in
favor of a network which dramatically speeds the propagation of items
when found.

In addition to the modest reduction in samples taken, the mailing-list
model has lower communication complexity than the original distributed
algorithm, generates less unproductive network traffic, and does not
assume the availability of global communication. The algorithm in
\cite{awerbuch05improved} communicates at every other step and has
communication complexity
$\ell\left(\frac{\eta}{\mbox{\scriptsize{min}}_i\{|P(S_i)|\}} +
\mu\ln(\mbox{max}_i\{|U(S_i)|\})\right)$. In contrast, the
mailing-list model only communicates when items of possible interest
are found, leading to a communication complexity of $$\ell \mbox{
  avg}_i\!\left(|S_i|\frac{\mbox{\scriptsize{avg}}_{u \in
    S_i}|P(u)|}{|P(S_i)|} \right)\mbox{.}$$ This is far less than the
distributed algorithm and is independent of the size of the set of
objects, which is particularly important when $|P(u)| \ll \eta$.

Of the traffic generated, a portion of it is ``spam'' from the
recipient's point of view. The spam generated in the mailing-list
model consists of personal interests broadcast, totalling $$\ell
\mbox{ avg}_i\!\left(|S_i|\frac{\mbox{\scriptsize{avg}}_{u \in
    S_i}|P(u)|}{|P(S_i)|} - 1\right)\mbox{.}$$ This could be reduced
further by extending the protocol to ``test the waters'' by checking
with a small sample of neighbors for each list prior to broadcasting
an item, but this would require the agents to be more aware of the
system's structure.  The original distributed algorithm is more
complicated to analyze in depth.  All queries prior to the discovery
of a SIG item are spam, and as the propagation of those items begins
many more queries will be sent to agents that do not hold items of
interest.  By itself, the former consists of $\ell
\frac{\eta}{\mbox{\scriptsize{min}}_i\{|P(S_i)|\}}$ messages, which by
itself will be more spam than the mailing-list model in many domains
(when $\eta \gg \mu$).

As noted earlier, we assume that SIG members have something meaningful
in common, i.e., $\frac{P(u)}{\eta} < \frac{P(S)}{P(v)}$.  However in
the mailing-list model is is also desirable for the ratio $|P(u)| /
|P(S)|$ to be $O(1)$, to prevent an inordinate number of broadcasts
from being made. This is in addition to the requirements of
\cite{awerbuch05improved}, in which the amount of superfluous traffic
is not a consideration.

The following network model removes the constraint on $|P(u)| /
|P(S)|$ and replaces the presence of cliques with the need for a
network in which subgroups of vertices are expected to be highly
connected (to the tune of an large fraction $\lambda$ of their
members) and which is an expander graph with a small expansion
coefficient.

\section{The Word-of-Mouth Model}\label{section-WMM}

In a random graph, it is highly unlikely that SIGs will be represented
by cliques.  Still, using a sample-and-share approach to spreading
information, we could achieve similar results to a mailing-list if
SIGs were sufficiently-connected subgraphs in $G$.  In random
scale-free networks it appears that the connected portions of SIGs
represent most of the members of the SIG under certain circumstances
($\beta_G$ significantly less than $\beta_0$, and $|S|$ a respectable
fraction of the population size $|U|$).  If most of a SIG's nodes
formed a connected component, either due to this result or by
construction\footnote{The results in Section~\ref{section-SFNs} are
  for SFNs created directly from a degree distribution, but SFNs can
  be formed by processes that exhibit growth and preferential
  attachment \cite{albert02statistical}.  One could imagine a network
  growing with \begin{em}interest-driven\end{em} preferential
    attachment.  Assortativity \cite{newman03mixing} may be most
    important side effect of this; for our purposes, biased failure
    modes would need to be studied in this context.}, it would be
appropriate to have an agent A's neighbors sample items A has liked
and spread those items further if they also like the item.  For ease
of analysis, we will assume that nodes remember what has been
recommended to them, and do not resample objects or recommend objects
redundantly over edges in the network.

If an item in $P(S)$ is found and propagated to $\lambda |S|$ or more
SIG members, it will also cause non-SIG members to test the item -- in
particular those users adjacent to the SIG in the network.  A SIG
member finding something in $P(S)$ would lead to $(1+\gamma)|S|$ total
users testing the announced item, where $\gamma$ is the expansion
coefficient of the graph. For ``Internet-like'' SFNs\footnote{In
  \cite{gkantsidis03conductance} these SFNs are made with a power-law
  degree distribution with $2 < \beta < 3$ but are altered to ensure
  there is a network ``core'' of minimum degree 3, to which all nodes
  of degree 1 or 2 are connected.  They are fully connected, while a
  random SFN with the same $\beta$ would almost surely have a large
  number of secondary components.} as in
\cite{gkantsidis03conductance} Gkantsidis et al.\ show the core of the
network has expansion properties, and that the second eigenvalue
$\lambda_2$ of a stochastic matrix corresponding to a random walk on
the graph is bounded as
$$1-\Omega\!\left(\frac{1}{\log n}\right) < \lambda_2 < 1 -
\Omega\!\left(\frac{1}{\log^2 n}\right)\mbox{.}$$ For large networks
this means the eigenvalue gap is not large, implying $\gamma$ is small
\cite{linial04expanders}.  As in the SIG model, false alarms are
possible, and while idiosyncratic interests should not propagate very
far $(1+\gamma)|S|$ can be used as a conservative estimate.  Taking
this result across all SIGs, the system recommendation complexity is
at worst $\ell (1+\gamma) |S| \frac{|P(u)|}{|P(S)|}$, making the total
sample complexity of the word-of-mouth model $$\ell
\left(\frac{\eta}{\mbox{\scriptsize{min}}_i\{|P(S_i)|\}} +
\mbox{avg}_i\!\left\{ (1+\gamma) |S_i|\frac{\mbox{\scriptsize{avg}}_{u
    \in S_i}|P(u)|}{|P(S_i)|}\right\} \right)\mbox{.}$$ This is
comparable to the mailing-list model in performance, with the caveat
that only some fraction of the population $\lambda |U|$ has
satisfaction guaranteed and more or less spam may be generated
depending on $\gamma$.

In an SFN, this algorithm will require as much communication as there
are edges within the SIG and at the boundary to the subgraph of
$S$. However, announcing personal interests will be only a fraction of
that work, and will vary depending on the popularity of the actual
item discovered. This hints at the presence
of \begin{em}equitability\end{em}; unpopular items will not be
  forwarded to large groups of users as they were in the mailing-list
  model, and popular items will satisfy more users for the system's
  trouble.  This also suggests the theoretical result based on SIGs
  may be a coarse approximation of the actual algorithm's performance,
  because SIGs are no longer strictly defined.  Further analysis is
  necessary.

\section{Conclusions and Future Work}\label{section-conclusions}

By explicitly considering the role of the network and limiting the
scope of communication in a distributed recommender system, the
mailing-list model and the word-of-mouth model both appear to do
better in terms of sampling, communication, and spam complexity than
Awerbuch's original work.  Although we are particularly interested in
the ability of agents to share recommendations to one another on
behalf of the users they represent, our work is applicable to a more
general context than DRSs.  It provides a formal basis for agents to
share information with only their nearest neighbors under certain
circumstances, with an understanding of when a large portion of the
interested agents will eventually receive that information.

This is part of ongoing work in distributed systems and the use of
agents that share information to enhance distributed learning.  We are
confirming the new SFN results in simulation, and we realize the
conductance properties of SFNs and the affects of assortativity need
to be more thoroughly studied.  The current conductance result depends
on the low conductance of an SFN when $2 < \beta < 3$, which
corresponds to the parameter space in which SIGs must represent a
large fraction of the entire population (these two results are related
-- in a graph with poor expansion properties, more bottlenecks exist
that could fail and fragment the graph).  In addition, for $2 < \beta
< \beta_0$ we do not know of a published result on the precise
fractional size of the giant component of the graph, and we are
preparing to publish our experimental plot of this -- it does not
appear to approach 1 in any reasonable limit as it does for $\beta <
2$.  This increases the necessary SIG size to accommodate $\lambda$,
and increases the contrast of our work to \cite{cohen00resilience}.

Both network models shown would benefit if agents could distinguish
their personal user's interests from those of each SIG or from those
of their neighbors, eliminating ``spam'' to the degree such
assessments were accurate.  Such information could also be used to
dynamically improve the graph structure in the word-of-mouth model, if
highly-correlated neighbors were introduced to one another.  If
features are added to the products of the current model, content-based
learning methods would be the next enabling step for this work.
Finally, we would like to identify systems that can share certain
kinds of information such that this sharing will lead independent
learners to converge on a common set of parameters.  This would make
data points the subject of recommendations, abstracted away from the
domain of the learned model.

\bibliographystyle{plain}
\bibliography{../bib/machine-learning,../bib/user-modeling,%
../bib/consensus-protocols,%
../bib/terran-pubs,../bib/hamilton-pubs,../bib/agents,../bib/mlrg}
\end{document}